\documentclass[10pt,a4paper]{article}
\usepackage[utf8]{inputenc}
\usepackage[T1]{fontenc}
\usepackage[a4paper,top=2.1cm,bottom=2.15cm,left=1.8cm,right=1.8cm,columnsep=0.68cm]{geometry}
\usepackage{microtype}
\usepackage{placeins}
\usepackage{dblfloatfix}
\usepackage{balance}
\usepackage{caption}
\usepackage{fancyhdr}
\usepackage{titlesec}
\usepackage{enumitem}
\titleformat{\section}{\normalsize\bfseries}{\thesection}{0.65em}{}
\titlespacing*{\section}{0pt}{1.4ex plus .4ex}{.6ex}
\titlespacing*{\subsection}{0pt}{1.1ex plus .3ex}{.4ex}
\usepackage{graphicx,booktabs,longtable,amsmath}
\usepackage[pdfencoding=auto,psdextra]{hyperref}
\hypersetup{pdftitle={Evidence from a Multi-Regime Causal Discovery Benchmark: UCMT Algorithms, Comparative Performance, and Controlled Ablation Studies},pdfauthor={Mahyar Eslami},pdfsubject={Causal discovery benchmark and component ablation study},pdfkeywords={causal discovery, UCMT, benchmark, ablation, directed F1, structural Hamming distance},hidelinks}
\usepackage[numbers]{natbib}
\title{Evidence from a Multi-Regime Causal Discovery Benchmark: UCMT Algorithms, Comparative Performance, and Controlled Ablation Studies}
\author{Mahyar Eslami\\\small UCMT Labs, Canada\\}
\date{8 November 2026 (revised 19 November 2026)\\\small Research manuscript (not peer-reviewed)}
\begin{document}
\emergencystretch=2em
\twocolumn[{\begin{@twocolumnfalse}\maketitle\vspace{-1.4em}\begin{abstract}
We present an internal, pre-registered evaluation of the UCMT causal discovery algorithm family alongside established alternatives across heterogeneous benchmark regimes, complemented by a separately pre-registered, controlled study of UCMT v2 Global components. The contribution is both empirical and methodological: a family-stratified comparison of three actively developed algorithm configurations and targeted evidence about which internal stages affect their outcomes. In the main benchmark, UCMT v2 Global has higher directed F1 than NOTEARS and DAGMA in selected real, semi-synthetic and Gaussian-network comparisons at the paired-run level, while trailing leading permutation-search methods in other settings. The BOSS-based UCMT v4 produces a CSuite-specific directed-F1 improvement over BOSS (mean paired difference +0.279; pair-level Holm $p<0.001$), but the dataset-level adjusted result is not significant ($p=0.188$). In a separate fixed-skeleton ablation, UCMT v2 Global exceeds an adapted R$^2$-SortnRegress orientation baseline on standardized iSCM data (directed F1 0.537 vs 0.468, dataset-level Holm $p=0.031$); absolute orientation accuracy is 0.563 and there is no established broad advantage over RESIT, ANM or DirectLiNGAM. Controlled ablations show that sparsity and BIC-based refinement materially affect some outcomes, whereas the examined arbitration and automatic regime-selection mechanisms do not yield confirmed general improvements. We report performance shortcomings, multiple-testing correction, runtime costs, evaluation limitations and reproducibility checks. The results support UCMT as an implemented and empirically investigated algorithm family, not a claim of overall superiority, independent validation or demonstrated commercial impact.
\end{abstract}

\paragraph{Competing interests.} This report was prepared by UCMT Labs, which develops the UCMT algorithms evaluated here. The author is the founder of UCMT Labs and the developer of the UCMT algorithms, and has a financial interest in their commercial use.

\paragraph{Evaluation status.} Both studies are internal, not peer-reviewed or independently replicated. The main benchmark and the later component experiments have different datasets, seed counts, methods and confirmatory families; their estimates are not pooled.
\paragraph{Intellectual property.} Only high-level algorithm descriptions and aggregate evaluation results are included. Detailed implementations and research settings remain confidential as trade secrets.
\vspace{0.65em}\end{@twocolumnfalse}}]

\section{Introduction}
Causal discovery from observational and interventional data is evaluated on few real datasets with trusted ground truth \citep{sachs2005,gamella2025} and on many synthetic generators whose artefacts can be exploited \citep{reisach2021}. We present an internal benchmark, pre-registered before execution, in which the UCMT family of algorithms is compared with established methods. The aim is to build a verifiable picture of where the UCMT family is useful, where it is not yet competitive, and which measured changes follow from its development. The public benchmark overview, methodology and comparative summaries are available at \url{https://ucmtlabs.com/benchmarks}. This page is a public results landing page, not a deposit of raw execution records or source code. No pooled ranking across families is produced. The paper offers three contributions: (i) a heterogeneous, family-specific comparison of v2 Global, v3 and a BOSS-based v4; (ii) separately pre-registered mechanism-level experiments on v2 Global; and (iii) an explicit translation of benchmark evidence into testable requirements for robust analytical software rather than unvalidated claims of business impact.

\section{Benchmark design}
\paragraph{Datasets.} Five families of part 1 (completed as far as the part-2 gap filling allowed) are analysed separately: real data (two Sachs versions and three Causal Chamber sets), semi-synthetic data (CausalAssembly stations and a SERGIO network \citep{gobler2024,dibaeinia2020}), DREAM4 \citep{marbach2010}, bnlearn Gaussian networks \citep{scutari2010} and CSuite \citep{geffner2022}. Part 2 adds three families, analysed separately and never pooled with the others: synthetic ER/SF (8 datasets), iSCM (6) and misspecification (4), chosen mechanically before the run, 3 seeds, 10 core methods (UCMT v2 Global, v3, v4, BOSS, GRaSP, PC, DirectLiNGAM, DAGMA, NOTEARS, empty graph); rankings there cover these 10 methods only. Discrete bnlearn and T\"ubingen families were not run for UCMT and are not interpreted.
\paragraph{Methods.} PC, FCI \citep{spirtes2000}, GES \citep{chickering2002}, GIES \citep{hauser2012}, GRaSP \citep{lam2022}, BOSS \citep{andrews2023}, LiNGAM-type methods \citep{shimizu2006,shimizu2011}, NOTEARS \citep{zheng2018}, DAGMA \citep{bello2022}, CAM \citep{buhlmann2014}, SCORE \citep{rolland2022}, DAS \citep{montagna2023das}, NoGAM \citep{montagna2023nogam}, AVICI \citep{lorch2022}, DiBS \citep{lorch2021}, SDCD \citep{nazaret2024}, DCDI \citep{brouillard2020}, ENCO \citep{lippe2022}, DECI \citep{geffner2022}, Sea \citep{wu2024}, plus Empty, random and sortability baselines \citep{reisach2021}.
\paragraph{UCMT algorithms (general description).} UCMT v2 Global is UCMT Labs' own algorithm, developed in-house; v3 is a variant of v2 Global with additional post-processing; v4 is a hybrid built on BOSS with an additional orientation step. The descriptions are general by design; the source code is confidential.
\paragraph{Protocol.} Pre-registration of settings, 10 seeds per dataset, standardised data, one thread per job, one-hour cap per job, UCMT dataset priors off, AVICI limited to 1,000 rows. Part 1 ended at its deadline, 2026-11-06 (it was interrupted on the host and its deadline moved by the time lost). Part 2 is a 24-hour pilot under Amendment 3 (pre-registered before any part-2 job) and used about 27 minutes of the budget; no job timed out.
\paragraph{Metrics and statistics.} SHD under a favourable convention (an undirected predicted edge on a true edge costs 0) and a strict convention (cost 1), directed F1, skeleton F1 and runtime. Ranks follow Dem\v{s}ar \citep{demsar2006} (Friedman and Nemenyi); paired comparisons use Wilcoxon signed-rank tests \citep{wilcoxon1945} with Holm correction \citep{holm1979}, and a dataset-level test where at least six datasets exist.

\section{Results}
\begin{table*}[!t]\centering\small
\caption{Datasets and methods ranked per family, and UCMT positions (favourable/strict SHD; DREAM4 directed F1) with the number of methods in the Nemenyi top group.}
\resizebox{\textwidth}{!}{\begin{tabular}{lrrlllr}\toprule
Family & Datasets & Methods & v2 Global & v3 & v4 & Top group \\ \midrule
Real data & 5 & 9 & 2/2 & 1/1 & 7/7 & 8 \\
Semi-synthetic & 6 & 17 & 4/1 & 6/2 & 2/4 & 15 \\
DREAM4 & 5 & 13 & 5/-- & 8/-- & 2/-- & 9 \\
bnlearn Gaussian & 8 & 17 & 5/5 & 7/6 & 1/2 & 9 \\
CSuite & 14 & 21 & 13/8 & 11/2 & 3/6 & 15 \\
Synthetic ER/SF & 8 & 10 & 6/6 & 5/5 & 2/1 & 7 \\
iSCM & 6 & 10 & 5/5 & 7/7 & 2/2 & 7 \\
Misspecification & 4 & 10 & 6/6 & 7/7 & 2/3 & 10 \\
\bottomrule\end{tabular}}\end{table*}
No method is separable from the first-ranked group in any family. On real data (five datasets) all UCMT variants have higher mean SHD than the empty graph (significant at pair level after Holm for UCMT v4 only) although their directed F1 (0.31 to 0.34) is far above it; with the four datasets of the earlier version all three were significant. On real-data favourable SHD UCMT v4 is significantly worse than (higher favourable SHD, pair-level Holm p < 0.05) UCMT v2 Global and UCMT v3 (50 pairs each), FCI, PC, SDCD and Sea (20 pairs each: these four ran on two of the five datasets only) and the Empty graph (50 pairs, Holm p 0.040).

\begin{table*}[!t]\centering\small
\caption{UCMT v4 minus BOSS on all common (dataset, seed) pairs: mean paired difference, wins/ties/losses, pair-level and dataset-level Holm $p$ (n/c: not computable).}
\resizebox{\textwidth}{!}{\begin{tabular}{llrrrr}\toprule
Family & Metric & Mean diff. & W/T/L & $p$ pair & $p$ dataset \\ \midrule
Real data & SHD strict & +0.280 & 8/31/11 & 1.000 & n/c \\
Real data & Dir.\ F1 & -0.002 & 9/28/13 & 1.000 & n/c \\
Semi-synthetic & SHD strict & -0.233 & 22/31/7 & 1.000 & 1.000 \\
Semi-synthetic & Dir.\ F1 & +0.018 & 23/23/14 & 0.254 & 1.000 \\
DREAM4 & Dir.\ F1 & -0.005 & 7/15/28 & $<$0.001 & n/c \\
bnlearn Gaussian & SHD strict & +0.037 & 7/65/8 & 1.000 & 1.000 \\
bnlearn Gaussian & Dir.\ F1 & -0.001 & 9/63/8 & 1.000 & 1.000 \\
CSuite & SHD strict & -1.300 & 75/65/0 & $<$0.001 & 0.207 \\
CSuite & Dir.\ F1 & +0.279 & 75/63/2 & $<$0.001 & 0.188 \\
\bottomrule\end{tabular}}\end{table*}
\begin{figure*}[!t]\centering\includegraphics[width=0.89\textwidth]{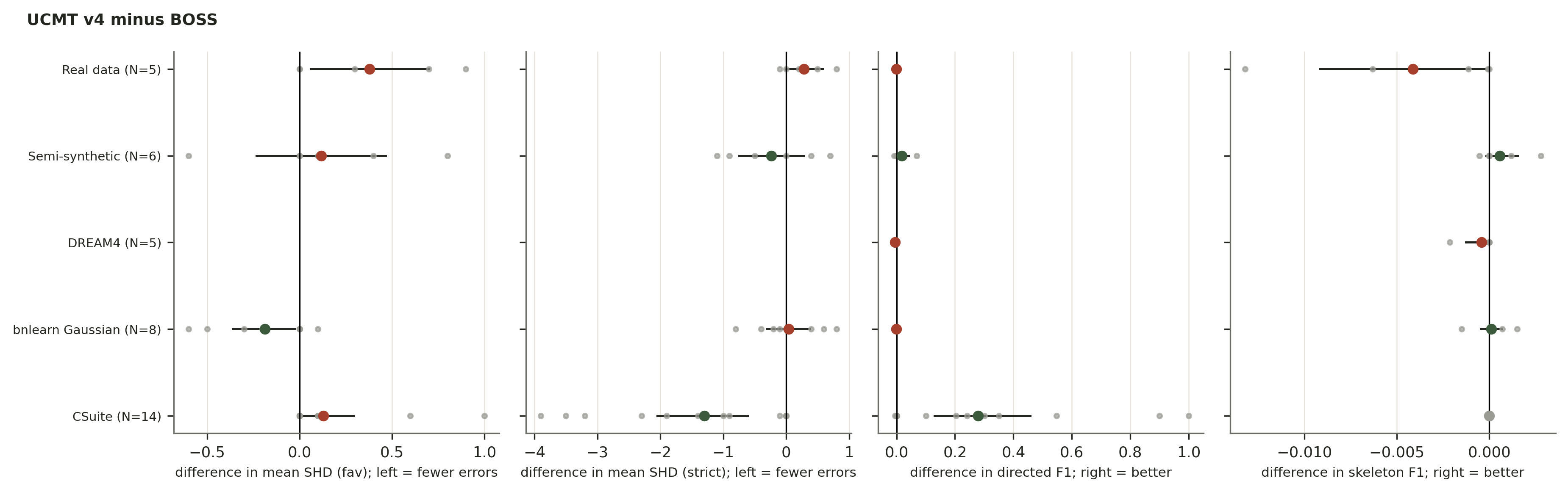}\caption{UCMT v4 minus BOSS: mean paired difference over datasets with 95\% bootstrap intervals.}\end{figure*}

\begin{table*}[!t]\centering\small
\caption{UCMT v2 Global versus BOSS, NOTEARS and DAGMA on directed F1: median paired difference (positive favours v2), wins/ties/losses and pair-level Holm $p$.}
\resizebox{\textwidth}{!}{\begin{tabular}{llrrr}\toprule
Family & Other & Median diff. & W/T/L & $p$ \\ \midrule
Real data & BOSS & -0.015 & 20/0/30 & 1.000 \\
Real data & NOTEARS & +0.134 & 50/0/0 & $<$0.001 \\
Real data & DAGMA & +0.110 & 49/0/1 & $<$0.001 \\
Semi-synthetic & BOSS & -0.021 & 25/3/32 & 1.000 \\
Semi-synthetic & NOTEARS & +0.127 & 54/3/3 & $<$0.001 \\
Semi-synthetic & DAGMA & +0.082 & 46/3/11 & $<$0.001 \\
DREAM4 & BOSS & -0.116 & 0/0/50 & $<$0.001 \\
DREAM4 & NOTEARS & +0.004 & 27/0/23 & 1.000 \\
DREAM4 & DAGMA & -0.023 & 15/0/35 & 0.311 \\
bnlearn Gaussian & BOSS & -0.108 & 22/0/58 & 0.010 \\
bnlearn Gaussian & NOTEARS & +0.238 & 58/0/22 & $<$0.001 \\
bnlearn Gaussian & DAGMA & +0.240 & 49/0/31 & $<$0.001 \\
CSuite & BOSS & +0.149 & 88/43/9 & $<$0.001 \\
CSuite & NOTEARS & +0.000 & 60/43/37 & $<$0.001 \\
CSuite & DAGMA & +0.000 & 62/23/55 & 1.000 \\
\bottomrule\end{tabular}}\end{table*}
\paragraph{Part 2 and cross-family robustness (Q5).} In the three part-2 families each UCMT variant is inside the Nemenyi top group of the ten core methods on favourable SHD, strict SHD and directed F1 (top group 7 to 10 of 10 methods; with four misspecification datasets the test cannot separate any methods). UCMT runtime grows more slowly with the number of variables than that of BOSS and GRaSP on the ER/SF subset (log-log slopes 1.8 against 2.8, D from 10 to 50). The pre-registered robustness question asked whether UCMT v2 Global is the most consistent fully directed method across families on directed F1, with a normalised mean rank r$_f$ of at most 0.50 in every part-2 family; it is not (r$_f$ = 0.59, 0.60 and 0.57), so the claim is unproven. Q5 was formulated after inspecting the part-1 results; on the part-1 families it is exploratory, and the confirmatory test is the pre-registered Q5 on the part-2 families. CPDAG-output methods are penalised by directed F1 for undirected edges, so the comparison is across output types; UCMT v4 (CPDAG output) has a lower worst-family rank than v2 Global in both parts, and on SHD v2 Global does not stand out (worst-family normalised rank W = 0.65 on both favourable and strict SHD).
\begin{table*}[!t]\centering\small
\caption{Q5, part-2 families, directed F1: normalised mean rank $r_f$ per family (lower is better), worst-family rank $W$ and bootstrap 95\% interval of $W$.}
\resizebox{\textwidth}{!}{\begin{tabular}{llrrrrr}\toprule
Method & Output & ER/SF & iSCM & Misspec. & $W$ & CI of $W$ \\ \midrule
GRaSP & CPDAG & 0.28 & 0.20 & 0.25 & 0.28 & 0.23--0.47 \\
UCMT v4 & CPDAG & 0.23 & 0.22 & 0.35 & 0.35 & 0.25--0.53 \\
BOSS & CPDAG & 0.26 & 0.18 & 0.38 & 0.38 & 0.25--0.53 \\
PC & CPDAG & 0.53 & 0.47 & 0.38 & 0.53 & 0.46--0.61 \\
UCMT v2 Global & DAG & 0.59 & 0.60 & 0.57 & 0.60 & 0.57--0.72 \\
UCMT v3 & DAG & 0.56 & 0.68 & 0.57 & 0.68 & 0.57--0.82 \\
NOTEARS & DAG & 0.75 & 0.67 & 0.70 & 0.75 & 0.70--0.80 \\
DirectLiNGAM & DAG & 0.47 & 0.80 & 0.57 & 0.80 & 0.72--0.87 \\
DAGMA & DAG & 0.84 & 0.68 & 0.82 & 0.84 & 0.79--0.90 \\
Empty (baseline) & DAG & 1.00 & 1.00 & 0.90 & 1.00 & 1.00--1.00 \\
\bottomrule\end{tabular}}\end{table*}
\begin{figure*}[!t]\centering\includegraphics[width=0.89\textwidth]{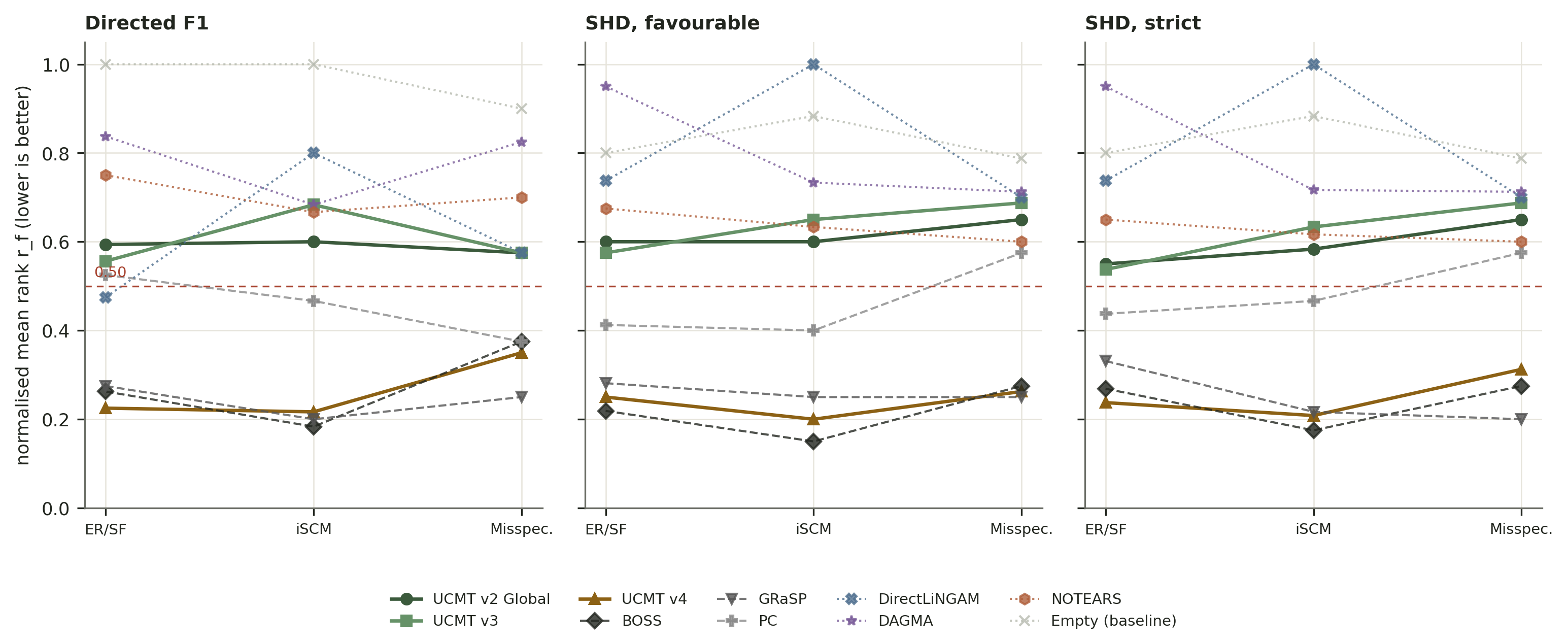}\caption{Normalised mean rank of the ten core methods in the part-2 families; dashed line: 0.50.}\end{figure*}
\begin{figure}[!t]\centering\includegraphics[width=0.98\columnwidth]{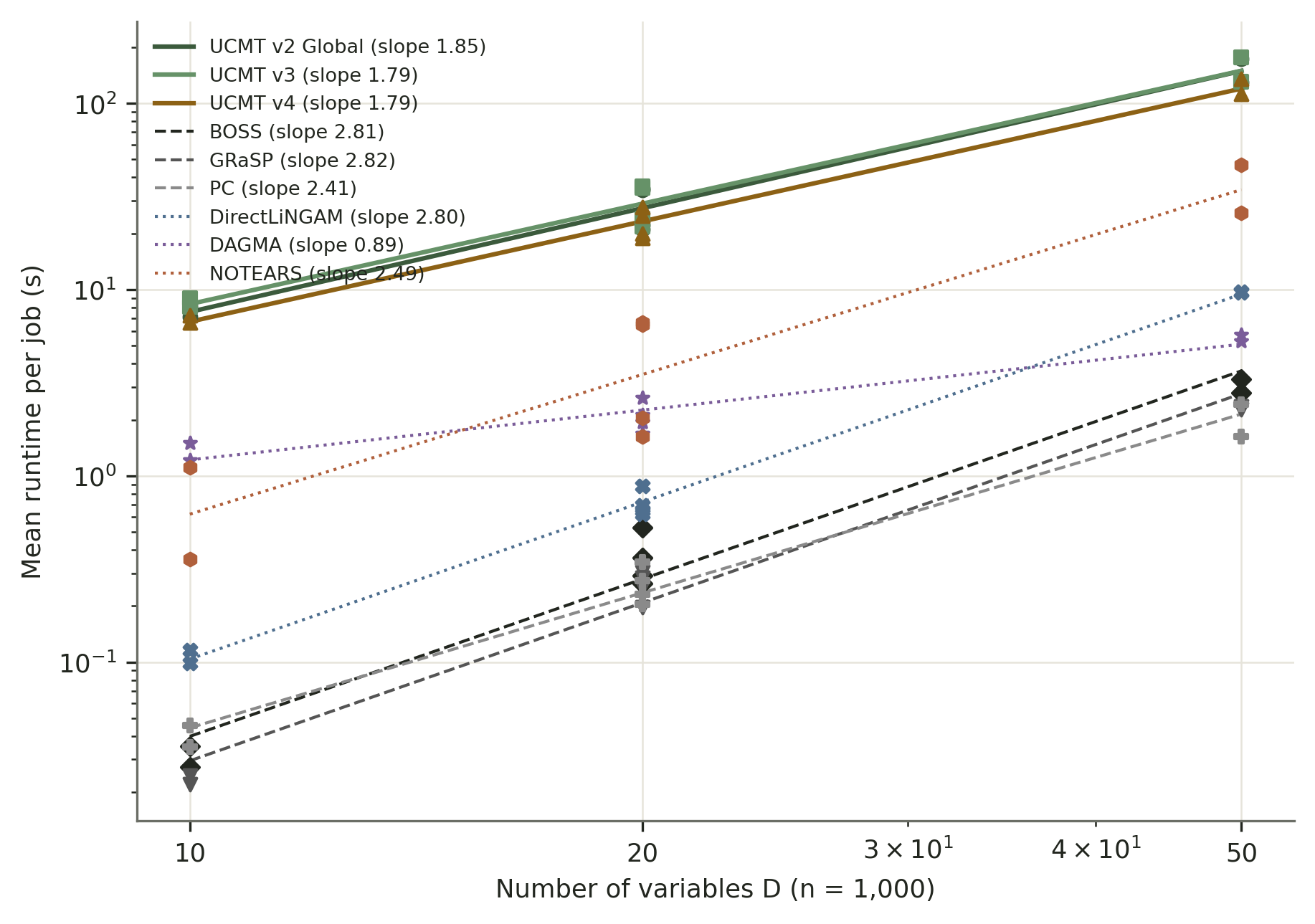}\caption{Runtime against number of variables on the ER/SF subset (log-log).}\end{figure}
\begin{figure*}[!t]\centering\includegraphics[width=0.89\textwidth]{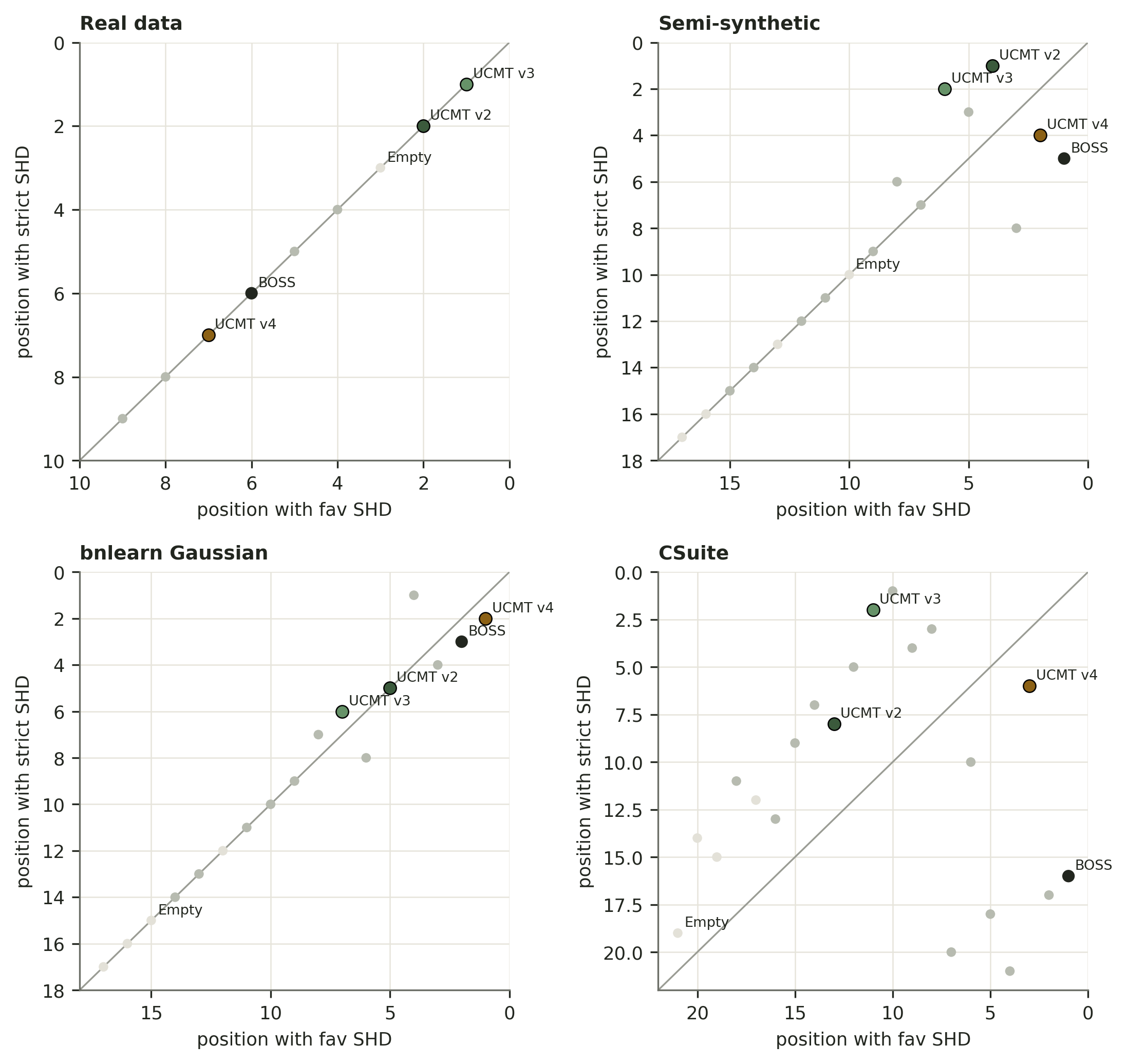}\caption{Sensitivity of method positions to the SHD convention.}\end{figure*}

\section{Controlled UCMT v2 Global component study}
\paragraph{Design and scope.} A second study was pre-registered on 2026-11-06 before the main-run outcomes were inspected and completed on 2026-11-07. It is distinct from the multi-algorithm benchmark above. It evaluated raw and standardized inputs using dataset-level paired analyses with two-sided Wilcoxon signed-rank tests, Holm adjustment within pre-specified confirmatory families, and two-stage bootstrap confidence intervals. The study logged 6,234 job records in the main and post-hoc queues plus 694 attempted CAM-GAM pruning comparisons. Among its evaluation groups were 12 classic synthetic, eight iSCM, eight misspecification, seven pooled real/Gaussian datasets, and four exploratory D=50 datasets. Dataset seeds are correlated within datasets and were not treated as independent confirmatory units. Evaluation used 10 seeds on smaller datasets and four or five at larger dimension. The primary validation compared an instrumented run with an unwrapped production run; no production source file was edited. This study does not represent independent replication.

\begin{table*}[!t]\centering\small
\caption{Summary of the separately pre-registered v2 component experiments. ``Confirmed'' refers only to the reported dataset-level, Holm-adjusted tests, not novel patentable subject matter.}
\begin{tabular}{p{0.65cm}p{4.1cm}p{6.9cm}}\toprule
Test & Evaluation question & Evidence-based outcome \\ \midrule
E1 & Combined orientation vs simpler orientation variants & No Holm-significant dataset-level advantage; ordering-only matched or exceeded the combined rule in tested groups. \\
E2 & v2 orientation on a fixed skeleton vs adapted published baselines & Positive narrow result on standardized iSCM vs adapted R$^2$-SortnRegress: +0.068 directed F1 and +0.094 orientation accuracy; both Holm $p=0.031$. \\
E3 & Automatic selection of data-processing regime & No confirmatory benefit; standardized inputs selected the hybrid branch in 346/346 cases. \\
E4 & Skeleton filtering variants vs PC-based skeletons & PC and rank-PC yielded higher skeleton F1 on classic synthetic data after correction; a small internal multi-path-filter gain was not Holm-significant. \\
E5 & Refinement stages and alternative pruning & Top-k sparsity control and BIC refinement improved SHD relative to absent/partial processing in some confirmatory synthetic comparisons; the nested F-test contributed little beyond top-k + BIC. \\
\bottomrule\end{tabular}
\end{table*}

\subsection{A bounded positive orientation result}
On eight standardized iSCM datasets (80 dataset-seed instances), UCMT v2's production orientation stage on the same true skeleton as its comparator achieved mean directed F1 of 0.537 versus 0.468 for the adapted R$^2$-SortnRegress order, a paired mean difference of 0.068 (95\% bootstrap interval 0.022--0.119). Orientation accuracy was 0.563 versus 0.468, a difference of 0.094 (95\% interval 0.045--0.147). The two dataset-level Holm-adjusted $p$-values were 0.031. This supports a narrowly specified comparative result on low-sortability iSCM data; it does not establish high absolute accuracy, robust causal orientation across regimes, or superiority to the broader set of baseline methods. All comparisons used adapted baselines constrained to the same skeleton and therefore do not compare published algorithms' native graph outputs.

\subsection{What the ablations actually establish}
The E5 results show that the end-to-end graph is sensitive to sparsity control and score-based post-processing. For example, on standardized iSCM instances, the full refinement chain had mean SHD 23.288 versus 40.700 without post-processing; this particular group-level comparison had dataset-level unadjusted $p=0.008$ but Holm-adjusted $p=0.383$ and must not be presented as confirmatory. Across the E5 confirmatory tests, seven standardized classic-synthetic comparisons survived dataset-level Holm adjustment; none did in raw data. Top-k plus BIC often matched the more elaborate chain. The results indicate engineering levers for simplification and calibration, not proof of a new pruning principle.

In E1 no tested combination of ordering and pairwise signals showed a confirmed dataset-level advantage after multiplicity correction. For E3, the automatic route was inactive on standardized inputs with respect to selection of the linear branch. E4 found that established PC-type methods can outperform the v2 intermediate skeleton in the classic-synthetic family. These negative and mixed findings are valuable because they identify components that can be simplified, replaced or investigated on held-out data before claims of improved performance are made.

\paragraph{Integrity and missing tests.} The research audit reports 17,987 independently reaggregated rows with no metric mismatches; the copied orientation rule and reference implementation matched, and the full E5 chain reproduced the production output in completed comparable runs. Limitations remain: no third-party reproduction, no new held-out dataset, no E4 filter-variant end-to-end comparison, and no evaluation of discrete, interventional or time-series settings in this component study. The existing v2 implementation also produced a ``math domain error'' on a subset of Causal Chamber \texttt{cc\_lt} runs (17 of 60 attempted runs across reported queues). Exclusions and comparator failures are documented in the internal execution audit.

\section{Research implications and deployment relevance}
\paragraph{Research contribution.} This work establishes a documented empirical baseline across model families and demonstrates that targeted mechanism tests can distinguish apparent overall performance from changes attributable to internal choices. The data identify a narrow, statistically supported orientation result and measurable contributions of refinement stages, while also challenging unvalidated assumptions about other components. This is an internally audited research process, not evidence of a universally improved causal discovery estimator.

\paragraph{Potential enterprise relevance (not yet validated).} Causal discovery tools may help analytical teams formulate hypotheses about operational drivers, prioritize follow-up tests and document causal assumptions. A deployable offering would additionally need validated uncertainty/abstention, stable data ingestion, task-specific benchmarks, interpretability, clear provenance, satisfactory compute cost and evidence from actual customer workflows. None of the benchmark results establishes an improvement in a customer's revenue, risk, decisions or operational outcomes. Those claims require prospective customer studies, explicit counterfactual or intervention-based validation where feasible, and agreed success metrics.

\paragraph{Practical development priorities.} The most directly supported next engineering tasks are (i) investigate the observed Causal Chamber numerical failures; (ii) test simpler orientation and post-processing configurations on held-out regimes; (iii) evaluate skeleton quality and final-graph quality separately; (iv) reduce absolute compute cost; and (v) obtain independent replication and prospective pilot evidence. They follow from the measured failure modes and comparative results rather than from a presumed performance lead.

\section{Training and pre-training disclosure}
Only AVICI and Sea use externally pre-trained networks; no UCMT component loaded trained weights in this benchmark, and dataset priors were off. Most methods fit a model or optimise at run time on each dataset's own data. No tuning loop on benchmark data was found for any method; for UCMT, whose development used Sachs and Causal Chamber data, the extent of tuning is not verified.
\begin{table}[!t]\centering\scriptsize
\caption{Primary kind of training for each algorithm and confidence of the evidence.}
\resizebox{\columnwidth}{!}{\begin{tabular}{llr}\toprule
Algorithm & Kind & Confidence \\ \midrule
UCMT v2 Global & Fitting at run time & High \\
UCMT v3 & Fitting at run time & High \\
UCMT v4 & Fitting at run time & High \\
PC & None & High \\
FCI & None & High \\
GES & None & High \\
GIES & None & High \\
GRaSP & None & High \\
BOSS & None & High \\
DirectLiNGAM & Fitting at run time & High \\
ICA-LiNGAM & Fitting at run time & High \\
DAGMA & Fitting at run time & High \\
NOTEARS & Fitting at run time & High \\
CAM (Python) & Fitting at run time & Medium \\
CAM (R) & Fitting at run time & Medium \\
SCORE & Fitting at run time & Medium \\
DAS & Fitting at run time & Medium \\
NoGAM & Fitting at run time & Medium \\
Var-SortnRegress & Fitting at run time & High \\
R2-SortnRegress & Fitting at run time & High \\
AVICI & Pre-training & High \\
Sea & Pre-training & High \\
DiBS & Fitting at run time & High \\
SDCD & Fitting at run time & High \\
DCDI & Fitting at run time & Medium \\
ENCO & Fitting at run time & High \\
DECI & Fitting at run time & High \\
Empty graph & None & High \\
RandomDAG & None & High \\
ANM / IGCI / RECI & Fitting at run time & Medium \\
\bottomrule\end{tabular}}\end{table}

\section{Discussion and limitations}
UCMT v2 Global is competitive with continuous-optimisation methods on directed F1 but not with permutation search on large Gaussian networks; UCMT v4 differs from BOSS only where its orientation step acts (about half of the undirected CSuite edges, almost none elsewhere); runtimes are 10 to over 1,000 times those of BOSS. Limitations: five datasets in the real and DREAM4 families, a part-2 pilot with 3 seeds, 10 methods and 4 to 8 datasets per family, non-independent seeds on real data, low test sensitivity of dataset-level tests, the favourable SHD convention, an inclusion rule that drops slow methods, a single host, and no ablations within the original benchmark itself. Separate subsequent v2 ablations are reported in Section 4; they must not be used to retroactively strengthen the original benchmark's confirmatory claims. The results do not support claims of leading performance.

\section{Reproducibility and data availability}
A public overview of the 2026.2 benchmark, including methodology and comparative summaries, is available at \url{https://ucmtlabs.com/benchmarks}. This web page is not a repository of raw trial records, the component-ablation records, or executable code. The original benchmark materials and the component-study protocol, execution scripts, raw records and generated tables are held internally. No public repository or DOI for those underlying materials is claimed. The proprietary UCMT engine is not released. Reproduction by independent parties may require a sealed executable/container and a confidential evaluation agreement. Publication of supplementary materials requires a separate security and trade-secret review. The previously frozen protocol checksum and execution details are retained internally.

\section{Conclusion}
The UCMT algorithm family has been implemented, evaluated on diverse benchmark families and subjected to targeted internal ablations. Specific strengths include competitive family-dependent performance and a confirmatory, narrow fixed-skeleton orientation advantage over an adapted R$^2$-ordering baseline on standardized iSCM data. The evidence also reveals important limitations: sensitivity to data regime, high absolute runtime relative to some strong baselines, mixed or absent component-level effects, and incomplete external generalization evidence. A balanced reading supports a serious R\&D program with concrete next validation steps, not broad superiority or commercial impact claims. An independent held-out study and customer-specific validation are necessary before making stronger scientific or deployment assertions.

\paragraph{Author contributions.} Mahyar Eslami is the sole author. He conceived the UCMT algorithm family (v2 Global, v3, v4) and designed and implemented its software; designed the benchmark and its pre-registered protocol and amendments; designed and ran the controlled component (ablation) study; ran all experiments; curated the data; carried out the statistical analysis; and wrote, reviewed and edited the manuscript. He takes full responsibility for the content of this manuscript.

\paragraph{Funding.} This work received no external funding.

\paragraph{Use of generative AI tools.} Generative AI writing assistants were used to help draft, edit and format parts of the text. The UCMT-specific algorithm implementations and modifications, experimental results, analyses, and conclusions are the author's own. The author reviewed and edited all AI-assisted text and is fully responsible for it. AI tools are not authors of this work.

\balance

\bibliographystyle{plainnat}
\bibliography{references}
\end{document}